\documentclass{article}

\usepackage[preprint]{neurips_2023}

\usepackage[utf8]{inputenc} 
\usepackage[T1]{fontenc}    

\usepackage{url}            
\usepackage{booktabs}       
\usepackage{amsmath, amsfonts, amssymb, amsthm}       

\usepackage{mathtools}
\usepackage{nicefrac}       
\usepackage{microtype}      
\usepackage{xcolor}         
\usepackage{graphicx}
\usepackage{multirow}       
\usepackage{subcaption}     
\usepackage{bbm}
\usepackage{algorithm, algorithmic}
\usepackage{placeins}
\usepackage{tabularx}
\usepackage{makecell}
\usepackage{wrapfig}
\theoremstyle{plain}
\newtheorem{theorem}{Theorem}[section]

\newtheorem{lemma}[theorem]{Lemma}

\theoremstyle{definition}
\newtheorem{definition}[theorem]{Definition}

\newcommand{\INDSTATE}[1][1]{\STATE\hspace{#1\algorithmicindent}}

\newenvironment{block}%
  {\list{}{\leftmargin=0.3in\rightmargin=0.3in}  \item[]  }%
  {\endlist}

\title{Randomized Quantization is All You Need for Differential Privacy in Federated Learning}

\author{
  Yeojoon Youn \\
  Georgia Institute of Technology\\
  \texttt{yjyoun92@gatech.edu} \\
  \And
  Zihao Hu \\
  Georgia Institute of Technology \\
  \texttt{zihaohu@gatech.edu} \\
  \And
  Juba Ziani \\
  Georgia Institute of Technology \\
  \texttt{jziani3@gatech.edu} \\
  \And
  Jacob Abernethy \\
  Google Research \\
  \& Georgia Institute of Technology \\
  \texttt{abernethyj@google.com} \\
}

\definecolor{red}{RGB}{255,00,00}

\usepackage{color}              
\usepackage{color-edits}
\addauthor{yeojoon}{cyan}
\addauthor{zihao}{blue}
\addauthor{juba}{red}
\addauthor{jake}{magenta}

\usepackage{hyperref}       
\usepackage{cleveref}
\begin{document}

\maketitle

\begin{abstract}
Federated learning (FL) is a common and practical framework for learning a machine model in a decentralized fashion. A primary motivation behind this decentralized approach is data privacy, ensuring that the learner never sees the data of each local source itself. Federated learning then comes with two majors challenges: one is handling potentially complex model updates between a server and a large number of data sources; the other is that de-centralization may, in fact, be insufficient for privacy, as the local updates themselves can reveal information about the sources' data. To address these issues, we consider an approach to federated learning that combines quantization and differential privacy. Absent privacy, Federated Learning often relies on quantization to reduce communication complexity. We build upon this approach and develop a new algorithm called the \textbf{R}andomized \textbf{Q}uantization \textbf{M}echanism (RQM), which obtains privacy through a two-levels of randomization. More precisely, we randomly sub-sample feasible quantization levels, then employ a randomized rounding procedure using these sub-sampled discrete levels. We are able to establish that our results preserve ``Renyi differential privacy'' (Renyi DP). We empirically study the performance of our algorithm and demonstrate that compared to previous work it yields improved privacy-accuracy trade-offs for DP federated learning. To the best of our knowledge, this is the first study that solely relies on randomized quantization without incorporating explicit discrete noise to achieve Renyi DP guarantees in Federated Learning systems.
\end{abstract}

\section{Introduction}

Federated Learning (FL) is an innovative approach to training on massive datasets, utilizing a multitude of devices like smartphones and IoT devices, each containing locally stored, privacy-sensitive data. At a basic level, privacy is maintained by storing local data on each end-user device without sharing it with the server. However, in some cases, device or local device data can be partially reconstructed from computed gradients~\citep{zhu2019deep}. This potential data leakage from gradients can be addressed through the use of privacy-preserving techniques. Equally important in the context of FL is communication efficiency. Given the extensive communication demands placed on many edge computing devices and the constraints of limited bandwidth, it is imperative to devise a training scheme that not only preserves privacy but also aligns with the requirements of efficient communication within FL systems. What is perhaps surprising, however, is that these two objectives are not necessarily in tension and \textit{can even be aligned!} One way to improve communication overhead is to reduce bit complexity through stochastic rounding schemes, but we show that these randomization procedures, if designed carefully, provide additional benefits to data privacy.

Past studies, such as the work of~\citet{agarwal2018cpsgd, kairouz2021distributed, agarwal2021skellam}, have sought to tackle this issue by employing various forms of discrete additive DP noise in conjunction with quantization. This is because the application of continuous noise post-quantization would effectively render the noise update continuous, negating any benefits to communication efficiency. However, when these discrete additive noise methods are coupled with secure aggregation protocols~\citep{bonawitz2017practical}, aimed at preventing a server from inspecting individual local device updates, they encounter a challenge of biased estimation due to gradient clipping. To solve this,~\citet{chen2022poisson} introduce the Poisson Binomial Mechanism (PBM), bypassing the use of additive noise and instead directly mapping continuous inputs to discrete values in an unbiased fashion.

Losses in accuracy compared to noise-free gradient updates that do not protect privacy in the strong sense afforded by differential privacy are essentially unavoidable. The addition of a privacy requirement inevitably constraints the learner's problem, and privacy must be traded-off with accuracy. Yet, the solution provided by~\citet{chen2022poisson}, while providing good performance, may still enjoy sub-optimal privacy-accuracy trade-offs. Can we develop new mechanisms with improved privacy-accuracy trade-off compared to the mechanism of~\citet{chen2022poisson}? 

Our starting point to address this question is to note that much research focusing on quantization in federated learning for the sake of communication complexity absent privacy~\citep{alistarh2017qsgd, reisizadeh2020fedpaq, haddadpour2021federated, youn2022accelerated} reveal that performance degradation from quantization alone is somewhat minimal. Furthermore, quantization itself inherently reduces the amount of information encoded about the original input.
While quantization in itself is insufficient for privacy, we posit that a two-stage approach, selecting a \emph{randomized} quantization scheme followed by randomized rounding, can provide a viable approach to obtaining low communication complexity, formal \emph{differential privacy} guarantees, while still enjoying good performance. Thus:
\begin{block}
    \emph{Can we harness randomization in quantization schemes to further improve privacy-accuracy trade-offs in differentially private federated learning?}
\end{block}

To address this question, we introduce what we call the \emph{Randomized Quantization Mechanism}, or \emph{RQM} for short. RQM achieves privacy entirely through randomly sub-sampling quantization levels followed by a (randomized) rounding procedure to a close-by quantization level. 

\paragraph{Summary of contributions} As mentioned above, our paper studies mechanisms for releasing gradients while satisfying Renyi differential privacy, and how our proposed mechanisms can be integrated in standard federated learning frameworks. 
\begin{itemize}
    \item In Section~\ref{sec:RQM}, we introduce our \emph{Randomized Quantization Mechanism} that maps gradients to a randomized discrete grid in a way that preserves Renyi differential privacy.
    \item In Section~\ref{sec:analysis}, we provide theoretical evidence that our proposed Randomized Quantization Mechanism exhibits $\alpha-$Renyi differential privacy guarantees ``locally'', at the level of each single end-user device. Our theoretical guarantees hold for $\alpha \to +\infty$, implying in particular that they hold not just for Renyi but also for traditional $(\epsilon,0)$-differential privacy.
    \item In Section~\ref{sec:experiments}, we provide experiments that highlight the performance of our mechanism. In particular, we show that RQM outperforms the state-of-the-art Poisson Binomial Mechanism (PBM) introduced by \citet{chen2022poisson} in two ways. First, we show that for any given $\alpha$, RQM provides lower Renyi divergence hence better Renyi DP guarantees than the work of~\citet{chen2022poisson}. Second, we also show that when RQM is plugged in the standard differentially private federated learning framework, it leads to high model accuracy when compared to that demonstrated by~\citet{chen2022poisson}.
\end{itemize}

\section{Related work}

Both communication complexity and privacy concerns have been driving forces behind the development of Federated Learning. Federated optimization often uses two types of privacy-preserving techniques hand-in-hand. One is secure multi-party computation, which protects the communication between local devices and the learner, preventing an attacker from intercepting messages sent between them~\citep{bell2020secure, bonawitz2017practical}. One is information-theoretic privacy guarantees such as differential privacy~\citep{dwork2014algorithmic} that prevent inference of any given single local device's data from observed summary outputs (such as local gradient updates or the learner's model itself). For example,~\citet{mcmahan2017learning} and \citet{geyer2017differentially} add a calibrated amount of Gaussian noise to the average of clipped local device updates based on the FedAvg \citep{mcmahan2017communication} algorithm.

In this paper, we focus on providing robust Renyi differential privacy guarantees in federated optimization while maintaining high communication efficiency and good accuracy. Previous methods have often used an approach based on quantization followed by the addition of discrete noise to achieve both differential privacy guarantees and low communication efficiency. \citet{agarwal2018cpsgd} introduces the first communication-efficient federated optimization algorithm with differential privacy by incorporating quantization with the binomial mechanism. \citet{kairouz2021distributed} and \citet{agarwal2021skellam} employ discrete Gaussian and Skellam mechanisms, respectively, in conjunction with quantization and secure aggregation for enhanced privacy. However, the above methods lead to biased estimation due to 
the necessity of modular clipping. To address this issue, \citet{chen2022poisson} and \citet{chaudhuri2022privacy} propose unbiased mechanisms with improved privacy-accuracy trade-offs. \citet{chen2022poisson} encodes local devices' gradients into a parameter of the binomial distribution, allowing their mechanism to generate a sample from this distribution without the need for additive discrete noise. In contrast, rather than using known privacy mechanisms, \citet{chaudhuri2022privacy} introduces the \emph{Minimum Variance Unbiased mechanism} (MVU) to enhance the privacy-utility trade-off by solving an optimization problem designed to minimize the output variance of the mechanism while adhering to local differential privacy and unbiasedness constraints. Despite their progress in improving the privacy-utility trade-off, these methods do not fully exploit the privacy advantages offered by randomized quantization itself.

Our research is not the first to leverage compression techniques to achieve both communication efficiency and provable privacy benefits without incorporating additive discrete noise \citep{li2019privacy, gandikota2021vqsgd}. \cite{li2019privacy} assume a Gaussian input vector distribution for their sketching algorithms to ensure differential privacy guarantees, which might not be strictly necessary. \citet{gandikota2021vqsgd} ultimately first quantize the gradients updates, then randomize the quantized gradients via differential private mechanisms such as randomized response or Rappor~\citep{erlingsson2014rappor}. However, and to the best of our knowledge, our Randomized Quantization Mechanism is the first investigation that exclusively utilizes randomization of the quantization itself to attain improved Renyi DP guarantees within Federated Learning frameworks.

\section{Preliminaries}

\subsection{Differential privacy}

Given that we employ the concept of differential privacy to demonstrate the privacy guarantees of our Randomized Quantization Mechanism, we first present the original definition of differential privacy.

\begin{definition} \label{definition:original_dp}
((Approximate) Differential Privacy \citep{dwork2006our}) For $\epsilon, \delta \geq 0$, a randomized mechanism $\mathcal{M}: \mathcal{D} \rightarrow \mathcal{R}$ satisfies $(\epsilon, \delta)$-differential privacy if for any neighbor dataset $D, D^\prime \in \mathcal{D}$ differing by the addition or removal of a single user's records, it holds that
\begin{gather*}
    \textrm{Pr}(\mathcal{M}(D) \in E) \leq e^\epsilon \cdot \textrm{Pr}(\mathcal{M}(D^\prime) \in E) + \delta
\end{gather*}
for all events $E \subset \mathcal{R}$.
\end{definition}


In this paper, we also consider a variant of standard differential privacy called \emph{Renyi Differential Privacy} (or \emph{Renyi DP}), introduced in the seminal work of~\citet{mironov2017renyi}. We develop mechanisms that guarantee Renyi DP and by extension traditional DP. The use of Renyi DP allows for tight privacy accounting throughout the training iterations. Renyi differential privacy relies on first understanding the notion of \emph{Renyi divergence}:

\begin{definition} \label{definition:renyi_div}
(Renyi Divergence \citep{renyi1961measures}) Let $P$ and $Q$ be probability distributions defined over $\mathcal{R}$. The Renyi divergence of order $\alpha > 1$ is defined as
\begin{gather*}
    D_\alpha (P||Q) := \frac{1}{\alpha-1} \log \mathbb{E}_{x \sim Q}\Big[\Big(\frac{P(x)}{Q(x)}\Big)^{\alpha}\Big].
\end{gather*}
\end{definition}

Then, Renyi differential privacy is defined as follows:
\begin{definition} \label{definition:renyi_dp}
(Renyi Differential Privacy \citep{mironov2017renyi}) A randomized mechanism $\mathcal{M} : \mathcal{D} \rightarrow \mathcal{R}$ satisfies $(\alpha, \epsilon)$-Renyi differential privacy if for any neighbor dataset $D, D^\prime \in \mathcal{D}$ it holds that
\begin{gather}
    D_\alpha (P_{\mathcal{M}(D)}||P_{\mathcal{M}(D^\prime)}) \leq \epsilon.
\end{gather}
\end{definition}

When $\alpha \to \infty$, $(\alpha, \epsilon)$-Renyi DP in fact recovers standard $(\epsilon, 0)$-DP. However, Renyi DP provides a finer-grained definition of privacy in that its guarantees can be tailored to the specific value of $\alpha$ and corresponding Renyi divergence that one considers. We now state a major property of the Renyi divergence that is useful to our theoretical analysis.


\begin{lemma} \label{lemma:monotonicity}
(Monotonicity) $D_\alpha$ is nondecreasing in $\alpha$. I.e., $D_\alpha(P||Q) \leq D_{\alpha^\prime}(P||Q)$ for all $1\leq \alpha \leq \alpha^\prime \leq \infty$.
\end{lemma}



\subsection{User-level privacy}


In the context of federated learning, we employ differential privacy to mask the contribution of any individual local device, making it challenging for a potential adversary to discern whether a local device's dataset was utilized in the training process. As such, we need to extend the traditional item-level definition of differential privacy (Definition \ref{definition:original_dp}) by redefining what we mean by neighboring datasets. In this context, two datasets are considered neighboring if one dataset can be created by changing any subset of data points of a single user from the other dataset. This user-level perspective is relatively standard and is the same as the one studied by~\citet{mcmahan2017learning} and~\citet{levy2021learning}.

\begin{definition} \label{definition:user_level_dp}
(User-level DP \citep{levy2021learning}) For $\epsilon, \delta \geq 0$, a randomized mechanism $\mathcal{M}: \mathcal{D} \rightarrow \mathcal{R}$ satisfies $(\epsilon, \delta)$-user level DP if for any neighbor dataset $D, D^\prime \in \mathcal{D}$ satisfying $\text{d}_{\text{user}}(D, D^\prime) \leq 1$, it holds that
\begin{gather*}
    \textrm{Pr}(\mathcal{M}(D) \in E) \leq e^\epsilon \cdot \textrm{Pr}(\mathcal{M}(D^\prime) \in E) + \delta
\end{gather*}
for all events $E \subset \mathcal{R}$, where $\text{d}_{\text{user}}$ is defined with $n$ users as
\begin{gather*}
    D = (D_1, \cdots, D_n), \textrm{ where } D_i = \{z_{i, 1}, \cdots, z_{i, m_i}\} \rightarrow \text{d}_{\text{user}}(D, D^\prime) := \sum_{i=1}^n 1\{ D_i \neq D_i^\prime \}
\end{gather*}
\end{definition}

\section{Model}\label{sec:model}

We consider a federated learning set-up comprised of three types of entities: there are $n$ end-user devices, one secure aggregator called \emph{SecAgg}, and one learner. The learner's goal is to learn a machine learning model, parameterized by a $f$-dimensional vector $w \in \mathbb{R}^f$, using the data on the devices through Stochastic Gradient Descent (SGD). However, the learner does not access the data from the devices directly, both for communication efficiency and privacy reasons. Rather, at each time step $t$: 
\begin{enumerate}
\item Each end-user device $i$ computes a clipped gradient $g^i_t \in [-c,c]^f$ locally using the data on that device only. This gradient is then encoded into an integer $z^i_t$. This integer can be seen as the index of a discrete level in a discretization of the space of potential gradients.
\item The secure aggregator receives one message $z^i_t$ from each device $i$, which encodes information about the gradient $g^i_t$ computed by $i$. The aggregator aggregates them into a single message $z_t = \sum_i z^i_t$.
\item The server decodes $z_t$, computes the corresponding gradient $\hat{g}_t$, and takes a gradient step $w_{t+1} \leftarrow w_t - \eta \hat{g}_t$.
\end{enumerate}

The traditional approach to federated learning releases gradients exactly; this approach is, however, i) inefficient from a communication complexity perspective and ii) vulnerable when it comes to privacy. We address i) by discretizing (or ``quantizing'') the space of possible values of the gradients to a grid of size $m$ per coordinate of the gradient; in turn, we require only $f \times \log m$ bits to represent a single update by a single device. Regarding ii), we note that it is well-understood that releasing exact gradients can lead to privacy violations in that the secure aggregator and the learner can recover information about device $i$'s dataset $D_i$ through the gradient itself. To address this issue, instead of releasing the gradient $g^i_t$ directly, device $i$ releases a noisy quantization level $z^i_t = RQM(g^i_t)$, where RQM is a Randomized Quantization Mechanism that must satisfy Renyi differential privacy. The entire setup is described formally in Algorithm~\ref{alg:distributed_dp_RQM}. 

What algorithm~\ref{alg:distributed_dp_RQM} describes is essentially the well-known, generic \emph{Differentially Private Stochastic Gradient Descent} approach to federated learning \citep{mcmahan2017learning}. The focus and novelty of our work, however, come from the design of the private mechanism RQM itself. We propose our new mechanism in Section~\ref{sec:RQM}, characterize its privacy guarantees in terms of Renyi differential privacy theoretically in Section~\ref{sec:analysis} and empirically in Section~\ref{sec:experiments}, and study the accuracy of the resulting model when using privately-released gradient updates to train it in Section~\ref{sec:experiments}.

\begin{algorithm}[!h]
\caption{Distributed DP-SGD with RQM} \label{alg:distributed_dp_RQM}
\begin{algorithmic}[1]
\STATE{\bfseries Input:} $N$ local devices, each local device dataset $D_i \in \mathcal{D}$ ($i=1, \cdots, N$), clipping threshold $c$, RQM parameters $(\Delta, m, q)$, server learning rate $\eta$, initial vector $w_0$, loss function $f(w, D)$
\FOR{$t = 0, \cdots, T-1$}
\STATE Server broadcasts $w_t$ to $n$ sampled local devices from total $N$ local devices;
\FOR{each local device $i$ in parallel}
\STATE $g_t^i \leftarrow \text{Clip}(\nabla f(w_t, D_i))$;
\STATE $z_t^i \leftarrow \text{RQM}(g_t^i)$;
\STATE send $z_t^i$ to the secure aggregator SecAgg.
\ENDFOR
\STATE SecAgg outputs $z_t = \sum_{i=1}^n z^i_t$;
\STATE server decodes $\hat{g}_t \leftarrow -(c+\Delta)+\frac{2z_t (c+\Delta)}{n(m-1)}$;
\STATE server finds $w_{t+1} \leftarrow w_t - \eta \hat{g}_t$.
\ENDFOR
\end{algorithmic}
\end{algorithm}

\section{The Randomized Quantization Mechanism}\label{sec:RQM}

In this section, we introduce our main building block for privacy in federated learning. This building block provides a mechanism for privately releasing a scalar aggregate statistic of a single user's data in the form of a new algorithm called the \emph{Randomized Quantization Mechanism} (RQM). We remark that when dealing with $f$-dimensional vectors instead, we apply our Randomized Quantization Mechanism independently to each vector coordinate.

We first formally present our RQM mechanism, outlined in Algorithm~\ref{alg:quantization_mechanism}. Since our mechanism relies on a discrete probability distribution to choose the quantization, we show how this probability distribution over the quantizations translates into a probability distribution over outcomes of our mechanism on any given input $x$; this distribution over outcomes is crucial to characterize the level of privacy obtained by our mechanism. Finally, we theoretically analyze the Renyi differential privacy guarantees of RQM by using this distribution over outcomes.

\subsection{Randomized Quantization Mechanism} \label{sec:RQM_explanation}

In this section, we assume that each user outputs a continuous scalar input $x$ computed from their data; this can be viewed as the simplest case of local updates.

Our RQM algorithm is then comprised of three key components: (1) enlarging the output range beyond the input range and setting up evenly spaced quantization bins, (2) sub-sampling realized quantization levels, and (3) performing a randomized rounding procedure on the \emph{sub-sampled} (and only those) discrete levels to map a value $x$ to a quantization level. Each of these steps is crucial in ensuring the Renyi DP guarantees of the RQM, as we describe below. Formally, in each step, we perform the following operations:
\begin{enumerate}
\item We establish the output range of our mechanism by first augmenting the size of the input range. We do so by adding $\Delta$ to the upper bound $c$ and subtracting $\Delta$ from the lower bound $-c$ on the input data. This augmentation of the range is necessary for privacy: if we use the same range for the output, the quantization output for the maximum input ($x=c$) would subsequently always be $c$ subsequently, leaking a lot of information about $x$. After this, we establish $m$ initial, evenly spaced quantization levels ($B(0), B(1), \cdots, B(m-1)$) within this output range, which will be potential outputs of our mechanism.
\item Instead of using the entire set of quantization levels from step (1), we randomly sub-sample feasible quantization levels. 
We do so by including each discrete level for quantization with a carefully chosen probability $q$. 
The randomization of the quantization levels is necessary for privacy; otherwise, a value of $x$ would always map to the fixed set of two quantization levels deterministically. This immediately breaks differential privacy. 
\item We perform quantization on the sub-sampled discrete levels (and these sub-sampled levels only), achieving both robust privacy and unbiased estimation. We identify the quantization bin that houses the input $x$ and perform randomized rounding on $x$ within this interval. The specific probabilities employed for randomized rounding can be reviewed in Algorithm \ref{alg:quantization_mechanism}.
\end{enumerate}

The randomized quantization and rounding procedures described above are also illustrated later on in Figure~\ref{fig:subsampling}.

\begin{algorithm}
\caption{Randomized Quantization Mechanism} \label{alg:quantization_mechanism}
\begin{algorithmic}[1]
\STATE{\bfseries Input:} $c>0, x \in [-c, c]$, extend the upper bound and lower bound by $\Delta$, the maximum number of quantization levels $m$, include a certain quantization level with probability $q$
\STATE Set $X^{\text{max}}$: $X^{\text{max}} = c+\Delta$, max and min value of quantization levels is respectively $X^{\text{max}}$, $-X^{\text{max}}$.
\STATE Quantization bins: $i=0, 1, \cdots, m-1 \rightarrow B(i)=-X^{\text{max}}+\frac{2iX^{\text{max}}}{m-1}$.
\STATE sub-sample feasible quantization levels: 

\INDSTATE Always include $B(0), B(m-1)$ \& $i=1, 2, \cdots, m-2 \rightarrow$ include $B(i)$ with probability $q$.
\INDSTATE sub-sampled indices of quantization levels $\rightarrow i_1(=0), i_2, \cdots, i_l(=m-1)$ 

\STATE Quantization step:

\INDSTATE Find $i_{j^*} (i_1 \leq i_{j^*} \leq i_l)$ that satisfies $x \in [B(i_{j^*}), B(i_{j^*+1})]$. 
\INDSTATE Do randomized rounding on $x$ in this interval.
\INDSTATE $$z=
\begin{cases}
i_{j^*+1},~\textrm{with probability } \frac{x-B(i_{j^*})}{B(i_{j^*+1})-B(i_{j^*})}\\
i_{j^*},~\textrm{ } \textrm{ } \textrm{ } \textrm{ } \textrm{ } \textrm{ o/w}
\end{cases}$$

\RETURN{$z$}
\end{algorithmic}
\end{algorithm}

\subsubsection{Flexibility of RQM hyperparameters} \label{sec:RQM_flexibility}

The hyperparameters within our RQM algorithm offer enhanced flexibility, allowing for a more nuanced 
hyperparameter optimization when compared to PBM. RQM has in fact three hyperparameters $\Delta, q, m$, while PBM has two hyperparameters $\theta, m$ (See Algorithm 2 in \citet{chen2022poisson}). At a fixed number of discrete levels $m$, i.e. at a fixed level of communication complexity, this allows us to search over a bigger space of output distributions of quantization levels than \citet{chen2022poisson} through the choice of $(q,\Delta)$.
In Section \ref{sec:fl_experiment}, we show that this leads to RQM achieving better privacy-accuracy trade-offs than PBM.

\subsection{Resulting discrete distribution of outcomes} \label{section:discrete_prob_dist_RQM}


Given an input $x$ and parameters $m, q, \Delta$, we can compute the discrete probability distribution of outputs $Q(x)$ of RQM over the set of potential quantization levels ${B(0), B(1), \cdots B(m-1)}$. This discrete probability distribution is given in Lemma \ref{lemma:discrete_prob_dist}:

\begin{lemma} \label{lemma:discrete_prob_dist}
Let $m \in \mathbb{N}$, and $q \in (0, 1)$ be parameters of Randomized Quantization Mechanism $Q$. Define evenly spaced $m$ quantization levels $B(0), \cdots, B(m-1)$ as in Algorithm \ref{alg:quantization_mechanism}. Let $j$ be the unique integer such that $x \in [B(j), B(j+1))$. The probability distribution of outcomes of the Randomized Quantization Mechanism is given by:  
\begin{align} \label{eq:discrete_prob_dist}
&\textrm{Pr}\Big(Q(x)=i\Big) \nonumber \\
&=\begin{cases}
(1-q)^{j-i}  \Big((1-q)^{m-j-2}\frac{B(m-1)-x}{B(m-1)-B(i)} + \sum\limits_{k=j+1}^{m-2} q(1-q)^{k-j-1} \frac{B(k)-x}{B(k)-B(i)}\Big),~i=0,\\
q(1-q)^{j-i}  \Big((1-q)^{m-j-2}\frac{B(m-1)-x}{B(m-1)-B(i)} + \sum\limits_{k=j+1}^{m-2} q(1-q)^{k-j-1} \frac{B(k)-x}{B(k)-B(i)}\Big),~0< i \leq j,\\
q(1-q)^{i-j-1}  \Big((1-q)^{j}\frac{x-B(0)}{B(i)-B(0)} + \sum\limits_{k=1}^{j} q(1-q)^{j-k} \frac{x-B(k)}{B(i)-B(k)}\Big),~j+1 \leq i < m-1,\\
(1-q)^{i-j-1}  \Big((1-q)^{j}\frac{x-B(0)}{B(i)-B(0)} + \sum\limits_{k=1}^{j} q(1-q)^{j-k} \frac{x-B(k)}{B(i)-B(k)}\Big),~i = m-1.
\end{cases}
\end{align}
\end{lemma}

The full proof of Lemma \ref{lemma:discrete_prob_dist} is provided in Appx.~\ref{appx:proof_discrete_prob_dist}. Equation (\ref{eq:discrete_prob_dist}) exhibits different cases for $\text{Pr}(Q(x)=i)$. These cases depend on i) how the $i$-th quantization level compares to $j$, where $j$ is defined to be such that $x \in [B(j),B(j+1))$ 
and ii) on the two special cases $i=0$ or $i=m-1$. The probabilities corresponding to these two extreme values of $i$ differ from the rest in that we always incorporate the $0$-th and ($m-1$)-th discrete level, which influences our probability calculations.

\begin{figure*}[!h]
    \centering
    \begin{subfigure}[b]{0.55\textwidth}
    \includegraphics[width=\textwidth]{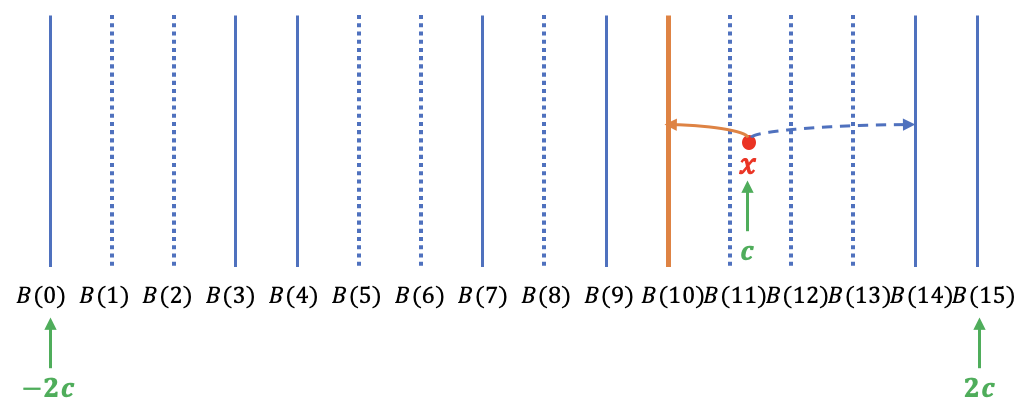}
    \vspace{2.5mm}
    \caption{An example of sub-sampling quantization levels for RQM.\vspace{2mm}}    \label{fig:subsampling}
    \end{subfigure}
    \quad
    \begin{subfigure}[b]{0.4\textwidth}
    \includegraphics[width=\textwidth]{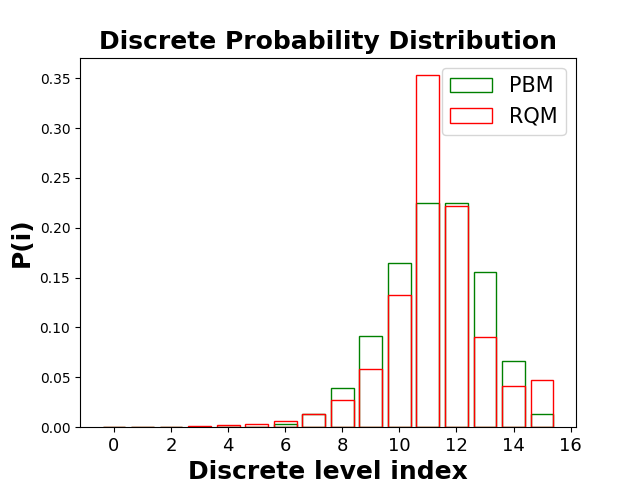}
        \caption{Distribution of outputs $Q(x)$ under PBM and RQM. 
        }  \label{fig:discrete_dist}
    \end{subfigure}
    \caption{An example of RQM with input $x=c$ and parameters $\Delta=c,~m=16$.}
\end{figure*}

Figures~\ref{fig:subsampling} and~\ref{fig:discrete_dist} provide some insights into how to derive the distribution of outputs and gives some intuition for Lemma \ref{lemma:discrete_prob_dist}. In Figure~\ref{fig:subsampling}, the solid lines corresponding to the ($0, 3, 4, 7, 9, 10, 14, 15$)-th discrete levels have been selected for quantization, while the dotted lines ($1, 2, 5, 6, 8, 11, 12, 13$)-th discrete levels have been thrown away. To have $Q(x) = 10$, the $10$-th discrete level must always be chosen while the $11$-th discrete level must not be chosen by the sub-sampling step of our algorithm; the probability of this happening is $q(1-q)$. The probability of the next quantization level bigger than $x$ being the $14$-th level, as shown in Figure~\ref{fig:subsampling}, is similarly given by $q (1-q)^2$ (levels $12$ and $13$ must not be sub-sampled, but $14$ must be). Then, the likelihood of $x$ transitioning to the $10$-th discrete level due to randomized rounding between the $10$-th and $14$-th levels is $\frac{B(14)-x}{B(14)-B(10)}$. I.e., the situation described in Figure~\ref{fig:subsampling} happens with probability $q^2 (1 - q)^3 \frac{B(14)-x}{B(14)-B(10)}$. For a complete analysis, we must also account for randomized rounding intervals $[B(10), B(12)], [B(10), B(13)], [B(10), B(15)]$ and aggregate all these probabilities. 

Figure~\ref{fig:discrete_dist} provides some insights on what the distribution induced by our quantization mechanism looks like, and seems to evidence that its shape differs from that of the Poisson Binomial Mechanism.

\subsection{Theoretical analysis of RQM's privacy guarantees}\label{sec:analysis}

We now provide a theoretical analysis of the level of Renyi-differential privacy achieved by our single-dimensional RQM mechanism. The full proof of Theorem \ref{theorem:single_device} is provided in Appx.~\ref{appx:proof_single_device}.
\begin{theorem} \label{theorem:single_device}
    Let $c, \Delta>0, \textrm{ } m \in \mathbb{N}$, and $q \in (0, 1)$ be parameters of Algorithm \ref{alg:quantization_mechanism}. Consider two scalars $x$ and $x'$ in $[-c,c]$, $P_{Q(x)}$ the distribution of outputs of RQM ran on scalar $x$, and $P_{Q(x')}$ the distribution of outputs of RQM ran on scalar $x'$. We have:
    \begin{gather}
        D_\alpha (P_{Q(x)}||P_{Q(x^\prime)}) \leq D_\infty (P_{Q(x)}||P_{Q(x^\prime)}) \leq \log \left(2(1-q)^2 \left(1 + \frac{c}{\Delta}\right)\right) + m\log\frac{1}{1-q}.
    \end{gather}
\end{theorem}

Our bound focuses on the case where $\alpha \to +\infty$, in which case $(\alpha,\epsilon)$-Renyi differential privacy is in fact the same as $(\epsilon,0)$-differential privacy as per the traditional definition of DP. There, we note that the privacy level $\epsilon = \log \left(2(1-q)^2 \left(1 + \frac{c}{\Delta}\right)\right) + m\log\frac{1}{1-q}$ that we obtain increases linearly on $m$, the number of quantizations level. This makes sense as a large number of quantization levels allows one to encode more information about the initial scalar $x$, in turn leading to less privacy and higher $\epsilon$'s. We also note that as $\Delta$ increases, $\epsilon$ decreases, and we obtain more privacy; once again, this follows the intuition from Section~\ref{sec:RQM_explanation} that when we increase the output range, we better protect the privacy of extreme values of $x$ that are close to $c$ or $-c$. As expected, when $\Delta = 0$, $\epsilon \to +\infty$ and our privacy guarantees are trivial, highlighting the fact that augmenting the range of output values beyond $[-c,c]$ is an unavoidable step to obtain reasonable privacy guarantees.

\section{Experiments}\label{sec:experiments}

In this section, we conduct experiments designed to complement our theoretical results and illustrate how RQM performs compared to PBM in terms of the privacy-accuracy trade-off. We first establish that RQM provides superior Renyi DP guarantees for the multiple-device scenario by numerically calculating the upper bound of Renyi divergence. Then, we employ the same parameters that yielded improved Renyi DP to demonstrate that RQM also excels in accuracy in our federated learning experiments.

\subsection{Numerical Renyi privacy guarantees}\label{sec:dp_experiment_result}

In Section~\ref{sec:analysis}, we characterized the privacy guarantee of our Randomized Quantization Mechanism in the special case in which $\alpha \to +\infty$. Our privacy guarantees hold at the \emph{local} level, in that it protects against a strong adversary that can see the output $Q(x_i)$ of each device $i$ (but not the input data $x_i$). A weaker but potentially interesting adversary can only see the output of the trusted, secure aggregator SecAgg\footnote{The learner only sees the output of SecAgg and is such an adversary.}. In this case, assuming we have $n$ devices that provide scalar inputs $x_1, \ldots, x_n$---where $x_i$ is the input of device $i$---to the randomized quantization mechanism RQM\footnote{We can think of these inputs as summary statistics computed on databases $D_1, \ldots, D_n$ before adding privacy; e.g., these could be a single coordinate of a clipped gradient.}, we are interested in an adversary that only sees the aggregate quantity $\sum_{i=1}^n Q(x_i)$. Given two vectors of inputs $x$ and $x^\prime$ that only differ in the input $x_i$ of a single device $i$, we obtain $(\alpha,\epsilon)$-Renyi differential privacy with
\[
\epsilon \triangleq D_\alpha \left(P_{\sum_{i=1}^n Q(x_i)}||  P_{\sum_{i=1}^n Q( x_i^\prime)}\right),
\]
as studied by~\citet{chen2022poisson}. 

\begin{figure*}[!htbp]
    \centering
    \begin{subfigure}[b]{0.4\textwidth}
    \includegraphics[width=\textwidth]{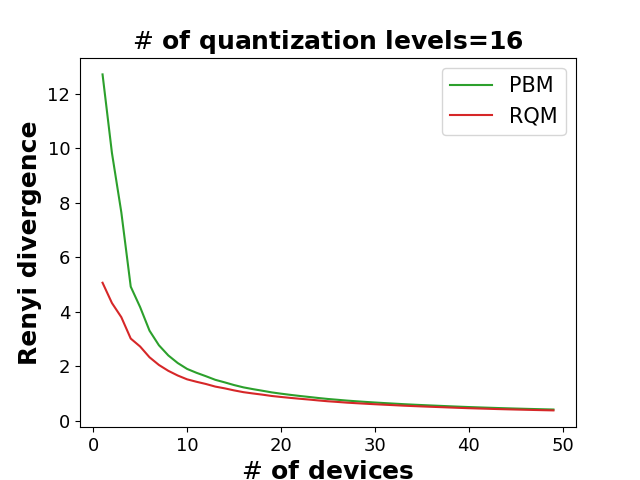}
    \end{subfigure}
    \quad
    \begin{subfigure}[b]{0.4\textwidth}
    \includegraphics[width=\textwidth]{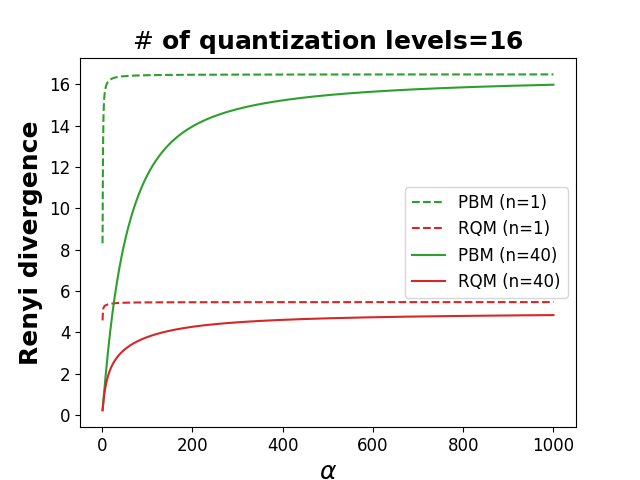}
    \end{subfigure}
    \caption{The left figure illustrates the inverse relationship between the number of devices $n$ and the upper bound of the Renyi divergence. The right figure indicates how the Renyi divergence increases as $\alpha$ increases.}
    \label{fig:renyi_devices_alpha}
\end{figure*}

We use Equation~\eqref{eq:discrete_prob_dist} to numerically compute and plot this Renyi divergence for finite $\alpha$ and for $n \geq 1$ in Figure~\ref{fig:renyi_devices_alpha}. We compare it to the Renyi divergence of the Poisson Binomial Mechanism of~\citet{chen2022poisson}; we note that we do not compare to the upper bound provided by~\citet{chen2022poisson} that may not be tight, but instead to the actual Renyi divergence computed numerically and \emph{exactly}. In both cases, we plot the \emph{worst-case} (over $x$, $x'$) Renyi divergence, which is maximized when all $x_i$'s are either $-c$ or $c$\footnote{This is because $(P, Q) \rightarrow D_\alpha(P||Q)$ is quasi-convex~\citep{van2014renyi}}; to do so, we generate $x$ and $x^\prime$ by taking $x_1=c, x_1^\prime=-c$ and randomly assigning a value of either $c$ or $-c$ to $x_2, \cdots, x_n$.

We fix the number of discrete levels $m$ as 16 for both RQM and PBM to compare privacy guarantees between the two algorithms \emph{at equal communication complexity}. We set the value $c$ to be $1.5$\footnote{Our mechanism is in fact scale-invariant for DP guarantees, and the choice of $c$ itself does not matter at a given constant ratio between $\Delta$ and $c$.}. Figure~\ref{fig:renyi_devices_alpha} is then obtained using the following parameters: $\theta=0.25$ for the hyperparameter of PBM\footnote{See Algorithm 2 in \citet{chen2022poisson} for a description of the parameters.}---we provide additional experiments comparing our results to PBM with different values of $\theta \in [0,0.4]$ in Appx.~\ref{appx:more_experiments}---and $\Delta=c$ with a corresponding fine-tuned $q = 0.42$ for RQM. The left picture fixes $\alpha = 2 $ and compares the Renyi divergences of PBM and RQM when $n$ increases. At the same time, the Renyi divergences seem to converge to similar values for $n \to +\infty$, and we note that our framework performs better for finite values of $n$, with a noticeably increasing performance gap as the number of user devices $n$ becomes smaller. The right picture fixes $n = 1$, then $n = 40$, and compares the Renyi divergence of PBM and RQM for a large range of $\alpha \in [0,1000]$; we see significant disparities in the levels of Renyi privacy guaranteed by PBM and RQM, with RQM vastly outperforming (i.e., guaranteeing a lower Renyi divergence hence a better privacy guarantee than) PBM, with the gap in privacy guarantees increasing as $\alpha \to +\infty$.


\subsection{Federated Learning experiments} \label{sec:fl_experiment}

In this section, we expand our experiments to provide insights beyond the privacy guarantees of the Randomized Quantization Mechanism itself, and to take into account how RQM integrates with the rest of the federated learning framework described in Section~\ref{sec:model}. We implement multi-dimensional RQM within the federated DP-SGD algorithm described in Algorithm \ref{alg:distributed_dp_RQM}). 

We evaluate the privacy-accuracy trade-off of our algorithms against the current leading approach, the Poisson Binomial Mechanism~\citep{chen2022poisson}, and an ideal noise-free clipped SGD benchmark that does not provide any differential privacy guarantee. The classification task for our federated learning experiment is performed on the EMNIST dataset~\citep{cohen2017emnist}. The experimental setup details are provided in Appx.~\ref{appx:experimental_setup_details}.

\paragraph{Hyperparameter choice.} We adhere to the same hyperparameters for our FL experiments as those of Section \ref{sec:dp_experiment_result}: $m=16$, $\theta=0.25$ for PBM, $\Delta=c, q=0.42$ for RQM. To highlight the flexibility of the choice of hyperparameters for RQM (Section \ref{sec:RQM_flexibility}), we also plot results of two more pairs $(\Delta, q) = (2c, 0.57)$ and $(\Delta, q) = (0.66c, 0.33)$. For clipping threshold $c$, we choose $2.9731\times 10^{-5}$. 

\begin{figure*}[!htbp]
    \centering
    \begin{subfigure}[b]{0.32\textwidth}
    \includegraphics[width=\textwidth]{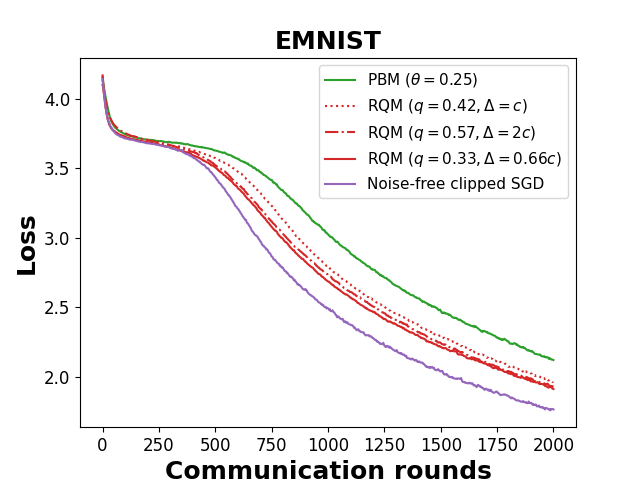}
    \end{subfigure}
    \quad
    \begin{subfigure}[b]{0.32\textwidth}
    \includegraphics[width=\textwidth]{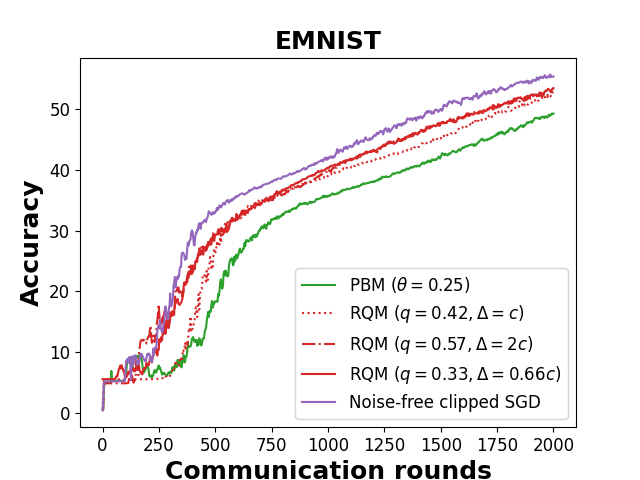}
    \end{subfigure}
    \begin{subfigure}[b]{0.32\textwidth}
    \includegraphics[width=\textwidth]{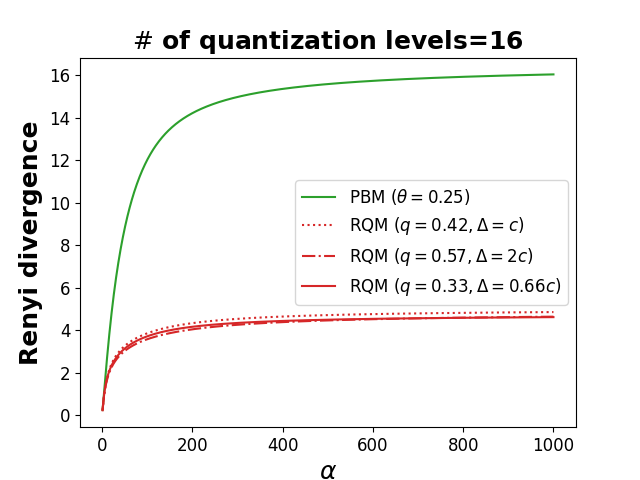}
    \end{subfigure}
    \caption{Comparing RQM with PBM and noise-free clipped SGD on EMNIST. All three RQMs with different hyperparameters outperform PBM in both a loss plot (Left) and an accuracy plot (Middle). These RQMs also show better Renyi DP guarantees than PBM (Right). }
    \label{fig:emnist_fl}
\end{figure*}

\paragraph{Experimental results} The left and middle plots in Figure \ref{fig:emnist_fl} clearly demonstrate that all three RQMs with different hyperparameter pairs show improved performance (in terms of loss and accuracy) on the EMNIST dataset than PBM. Among three RQMs, $(\Delta, q) = (0.66c, 0.33)$ achieves the best accuracy. The performance of our three RQMs are still worse than noise-free clipped SGD: this is unavoidable because noise-free clipped SGD only focus on accuracy without providing any privacy guarantees, and is an ideal, impossible-to-achieve benchmark with privacy. 

The right plot in Figure~\ref{fig:emnist_fl} replicates experiment of Section \ref{sec:dp_experiment_result} that were aimed at showcasing the privacy level achieved by RQM compared to PBM. The figure shows that the improved accuracy of the three RQMs compared to PBM in the left and middle figures does not come at the cost of privacy. In fact, the three plots together demonstrate that all three instantiations of RQM provide both better performance and better Renyi DP guarantees than PBM. I.e., in our experiments, RQM improves the \emph{privacy-accuracy trade-off} of federated differentially private stochastic gradient descent compared to the current state of the art.

\begin{ack}
This work was made possible through the support of the AI4OPT Institute under the NSF Award 2112533, and the support of the NSF Award IIS-1910077.
\end{ack}



\bibliography{neurips_2023}

\begin{thebibliography}{}

\bibitem[Abadi et~al., 2015]{tensorflow2015-whitepaper}
Abadi, M., Agarwal, A., Barham, P., Brevdo, E., Chen, Z., Citro, C., Corrado,
  G.~S., Davis, A., Dean, J., Devin, M., Ghemawat, S., Goodfellow, I., Harp,
  A., Irving, G., Isard, M., Jia, Y., Jozefowicz, R., Kaiser, L., Kudlur, M.,
  Levenberg, J., Man\'{e}, D., Monga, R., Moore, S., Murray, D., Olah, C.,
  Schuster, M., Shlens, J., Steiner, B., Sutskever, I., Talwar, K., Tucker, P.,
  Vanhoucke, V., Vasudevan, V., Vi\'{e}gas, F., Vinyals, O., Warden, P.,
  Wattenberg, M., Wicke, M., Yu, Y., and Zheng, X. (2015).
\newblock {TensorFlow}: Large-scale machine learning on heterogeneous systems.
\newblock Software available from tensorflow.org.

\bibitem[Agarwal et~al., 2021]{agarwal2021skellam}
Agarwal, N., Kairouz, P., and Liu, Z. (2021).
\newblock The skellam mechanism for differentially private federated learning.
\newblock {\em Advances in Neural Information Processing Systems},
  34:5052--5064.

\bibitem[Agarwal et~al., 2018]{agarwal2018cpsgd}
Agarwal, N., Suresh, A.~T., Yu, F. X.~X., Kumar, S., and McMahan, B. (2018).
\newblock cpsgd: Communication-efficient and differentially-private distributed
  sgd.
\newblock {\em Advances in Neural Information Processing Systems}, 31.

\bibitem[Alistarh et~al., 2017]{alistarh2017qsgd}
Alistarh, D., Grubic, D., Li, J., Tomioka, R., and Vojnovic, M. (2017).
\newblock Qsgd: Communication-efficient sgd via gradient quantization and
  encoding.
\newblock {\em Advances in neural information processing systems}, 30.

\bibitem[Bell et~al., 2020]{bell2020secure}
Bell, J.~H., Bonawitz, K.~A., Gasc{\'o}n, A., Lepoint, T., and Raykova, M.
  (2020).
\newblock Secure single-server aggregation with (poly) logarithmic overhead.
\newblock In {\em Proceedings of the 2020 ACM SIGSAC Conference on Computer and
  Communications Security}, pages 1253--1269.

\bibitem[Bonawitz et~al., 2017]{bonawitz2017practical}
Bonawitz, K., Ivanov, V., Kreuter, B., Marcedone, A., McMahan, H.~B., Patel,
  S., Ramage, D., Segal, A., and Seth, K. (2017).
\newblock Practical secure aggregation for privacy-preserving machine learning.
\newblock In {\em proceedings of the 2017 ACM SIGSAC Conference on Computer and
  Communications Security}, pages 1175--1191.

\bibitem[Chaudhuri et~al., 2022]{chaudhuri2022privacy}
Chaudhuri, K., Guo, C., and Rabbat, M. (2022).
\newblock Privacy-aware compression for federated data analysis.
\newblock In {\em Uncertainty in Artificial Intelligence}, pages 296--306.
  PMLR.

\bibitem[Chen et~al., 2022]{chen2022poisson}
Chen, W.-N., Ozgur, A., and Kairouz, P. (2022).
\newblock The poisson binomial mechanism for unbiased federated learning with
  secure aggregation.
\newblock In {\em International Conference on Machine Learning}, pages
  3490--3506. PMLR.

\bibitem[Cohen et~al., 2017]{cohen2017emnist}
Cohen, G., Afshar, S., Tapson, J., and Van~Schaik, A. (2017).
\newblock Emnist: Extending mnist to handwritten letters.
\newblock In {\em 2017 international joint conference on neural networks
  (IJCNN)}, pages 2921--2926. IEEE.

\bibitem[Dwork et~al., 2006]{dwork2006our}
Dwork, C., Kenthapadi, K., McSherry, F., Mironov, I., and Naor, M. (2006).
\newblock Our data, ourselves: Privacy via distributed noise generation.
\newblock In {\em Advances in Cryptology-EUROCRYPT 2006: 24th Annual
  International Conference on the Theory and Applications of Cryptographic
  Techniques, St. Petersburg, Russia, May 28-June 1, 2006. Proceedings 25},
  pages 486--503. Springer.

\bibitem[Dwork et~al., 2014]{dwork2014algorithmic}
Dwork, C., Roth, A., et~al. (2014).
\newblock The algorithmic foundations of differential privacy.
\newblock {\em Foundations and Trends{\textregistered} in Theoretical Computer
  Science}, 9(3--4):211--407.

\bibitem[Erlingsson et~al., 2014]{erlingsson2014rappor}
Erlingsson, {\'U}., Pihur, V., and Korolova, A. (2014).
\newblock Rappor: Randomized aggregatable privacy-preserving ordinal response.
\newblock In {\em Proceedings of the 2014 ACM SIGSAC conference on computer and
  communications security}, pages 1054--1067.

\bibitem[Gandikota et~al., 2021]{gandikota2021vqsgd}
Gandikota, V., Kane, D., Maity, R.~K., and Mazumdar, A. (2021).
\newblock vqsgd: Vector quantized stochastic gradient descent.
\newblock In {\em International Conference on Artificial Intelligence and
  Statistics}, pages 2197--2205. PMLR.

\bibitem[Geyer et~al., 2017]{geyer2017differentially}
Geyer, R.~C., Klein, T., and Nabi, M. (2017).
\newblock Differentially private federated learning: A client level
  perspective.
\newblock {\em arXiv preprint arXiv:1712.07557}.

\bibitem[Haddadpour et~al., 2021]{haddadpour2021federated}
Haddadpour, F., Kamani, M.~M., Mokhtari, A., and Mahdavi, M. (2021).
\newblock Federated learning with compression: Unified analysis and sharp
  guarantees.
\newblock In {\em International Conference on Artificial Intelligence and
  Statistics}, pages 2350--2358. PMLR.

\bibitem[Kairouz et~al., 2021]{kairouz2021distributed}
Kairouz, P., Liu, Z., and Steinke, T. (2021).
\newblock The distributed discrete gaussian mechanism for federated learning
  with secure aggregation.
\newblock In {\em International Conference on Machine Learning}, pages
  5201--5212. PMLR.

\bibitem[Levy et~al., 2021]{levy2021learning}
Levy, D., Sun, Z., Amin, K., Kale, S., Kulesza, A., Mohri, M., and Suresh,
  A.~T. (2021).
\newblock Learning with user-level privacy.
\newblock {\em Advances in Neural Information Processing Systems}, 34.

\bibitem[Li et~al., 2019]{li2019privacy}
Li, T., Liu, Z., Sekar, V., and Smith, V. (2019).
\newblock Privacy for free: Communication-efficient learning with differential
  privacy using sketches.
\newblock {\em arXiv preprint arXiv:1911.00972}.

\bibitem[McMahan et~al., 2017a]{mcmahan2017communication}
McMahan, B., Moore, E., Ramage, D., Hampson, S., and y~Arcas, B.~A. (2017a).
\newblock Communication-efficient learning of deep networks from decentralized
  data.
\newblock In {\em Artificial intelligence and statistics}, pages 1273--1282.
  PMLR.

\bibitem[McMahan et~al., 2017b]{mcmahan2017learning}
McMahan, H.~B., Ramage, D., Talwar, K., and Zhang, L. (2017b).
\newblock Learning differentially private recurrent language models.
\newblock {\em arXiv preprint arXiv:1710.06963}.

\bibitem[Mironov, 2017]{mironov2017renyi}
Mironov, I. (2017).
\newblock R{\'e}nyi differential privacy.
\newblock In {\em 2017 IEEE 30th computer security foundations symposium
  (CSF)}, pages 263--275. IEEE.

\bibitem[Reisizadeh et~al., 2020]{reisizadeh2020fedpaq}
Reisizadeh, A., Mokhtari, A., Hassani, H., Jadbabaie, A., and Pedarsani, R.
  (2020).
\newblock Fedpaq: A communication-efficient federated learning method with
  periodic averaging and quantization.
\newblock In {\em International Conference on Artificial Intelligence and
  Statistics}, pages 2021--2031. PMLR.

\bibitem[R{\'e}nyi, 1961]{renyi1961measures}
R{\'e}nyi, A. (1961).
\newblock On measures of entropy and information.
\newblock In {\em Proceedings of the Fourth Berkeley Symposium on Mathematical
  Statistics and Probability, Volume 1: Contributions to the Theory of
  Statistics}, volume~4, pages 547--562. University of California Press.

\bibitem[Van~Erven and Harremos, 2014]{van2014renyi}
Van~Erven, T. and Harremos, P. (2014).
\newblock R{\'e}nyi divergence and kullback-leibler divergence.
\newblock {\em IEEE Transactions on Information Theory}, 60(7):3797--3820.

\bibitem[Youn et~al., 2022]{youn2022accelerated}
Youn, Y., Kumar, B., and Abernethy, J. (2022).
\newblock Accelerated federated optimization with quantization.
\newblock In {\em Workshop on Federated Learning: Recent Advances and New
  Challenges (in Conjunction with NeurIPS 2022)}.

\bibitem[Zhu et~al., 2019]{zhu2019deep}
Zhu, L., Liu, Z., and Han, S. (2019).
\newblock Deep leakage from gradients.
\newblock {\em Advances in neural information processing systems}, 32.

\end{thebibliography}
\bibliographystyle{apalike}

\clearpage
\appendix

\section{Discussion}

In conclusion, this paper introduces a novel algorithm, the Randomized Quantization Mechanism (RQM). The RQM achieves privacy through a two-tiered process of randomization, which includes (1) the random subsampling of viable quantization levels, and (2) the application of a randomized rounding process with these subsampled discrete levels. We have theoretically demonstrated the Renyi differential privacy guarantees of RQM for a single end-user device and provided empirical evidence of its superior performance in the \emph{privacy-accuracy trade-off} compared to the state-of-the-art Poisson Binomial Mechanism (PBM). In the future, it would be worthwhile to further examine the Renyi DP guarantees of RQM for multiple-device scenarios and the case of multi-dimensional RQM from a theoretical standpoint. Furthermore, increasing the flexibility of RQM hyperparameters by assigning unique probability values $q_i$ to each $i$-th discrete level presents an intriguing avenue for further enhancing the privacy-accuracy trade-off.

\section{Missing proofs}

\subsection{Proof of Lemma \ref{lemma:discrete_prob_dist}} \label{appx:proof_discrete_prob_dist}

In Lemma \ref{lemma:discrete_prob_dist}, $j$ is defined as $x \in [B(j), B(j+1))$. We divide the range of $i$ into four cases-$0 <i \leq j$, $i=0$, $j+1 \leq i < m-1$, $i=m-1$- and compute the discrete probability $\text{Pr}(Q(x)=i)$ for each case. The core proof idea of Lemma \ref{lemma:discrete_prob_dist} is centered on evaluating the probability of each potential interval that can be used for randomized rounding for $x$. Subsequently, the probability that $Q(x) = i$ arises due to randomized rounding within a given interval is computed. Thus, when $i \leq j$ and $k \geq j+1$, we define the event $E_i$ and $F_k$ as below to use this notation for calculating the probability of each potential interval that can be used for the randomized rounding step in Algorithm \ref{alg:quantization_mechanism}.
\begin{align} \label{event_E_F_definition}
    E_i&: \text{ the event of }i\text{-th discrete level being used for randomized rounding} \nonumber \\
    F_k&: \text{ the event of }k\text{-th discrete level being used for randomized rounding}
\end{align}

From the above definition of two events, $E_i \cap F_k$ indicates an event of the interval $[B(i), B(k)]$ being used for randomized rounding. In this event, this also means $i_{j^*}=i$ and $i_{j^*+1}=k$ in Algorithm \ref{alg:quantization_mechanism}. Now, let's deep dive into how we can exactly calculate $\text{Pr}(Q(x)=i)$ for each case of four ranges.

(I) $0 < i \leq j$:

First, Let us consider the case when $0 < i \leq j$. Similar to the logic in Section \ref{section:discrete_prob_dist_RQM}, to have $Q(x) = i$, the $i$-th discrete level must always be chosen while the $(i+1)$-th, $\cdots$, $j$-th discrete levels must not be chosen by the sub-sampling step of our algorithm. The probability of this happening is $q(1-q)^{j-i}$. Thus, we can use the definition of the event $E_i$ in (\ref{event_E_F_definition}) for this case.
\begin{gather} \label{eq:i_chosen}
    \text{Pr}(E_i) = \text{Pr}(i: \text{ chosen}, \text{ }(i+1, \cdots, j): \text{ not chosen}) = q(1-q)^{j-i}
\end{gather}
Let us denote $k$ as an index of the next quantization level bigger than $x$. The possible $k$s are $j+1, \cdots, m-1$. When $k \in [j+1, m-2]$, the probability of the next quantization level bigger than $x$ being the $k$-th level is similarly given by $q(1-q)^{k-j-1}$. Thus, we can use the definition of the event $F_k$ in (\ref{event_E_F_definition}) for this case.
\begin{gather} \label{eq:k_chosen}
    \text{Pr}(F_k) = \text{Pr}(k: \text{ chosen}, \text{ }(j+1, \cdots, k-1): \text{ not chosen}) = q(1-q)^{k-j-1}
\end{gather}
Then, the likelihood of $x$ transitioning to the $i$-th discrete level due to the randomized rounding between $i$-th and $k$-th levels is $\frac{B(k)-x}{B(k)-B(i)}$. This means
\begin{gather} \label{eq:randomized_rounding}
    \text{Pr}(Q(x)=i|E_i \cap F_k) = \frac{B(k)-x}{B(k)-B(i)}
\end{gather}
Therefore, for $k \in [j+1, m-2]$, by combining (\ref{eq:i_chosen}), (\ref{eq:k_chosen}), (\ref{eq:randomized_rounding}), we get
\begin{align} \label{eq:quantization_i_before_m-1}
    &\text{Pr}((Q(x)=i) \cap E_i \cap F_k) \nonumber \\
    =& \text{Pr}(E_i \cap F_k) \cdot \text{Pr}(Q(x)=i|E_i \cap F_k) \nonumber \\
    =&  \text{Pr}(E_i) \cdot \text{Pr}(F_k) \cdot \text{Pr}(Q(x)=i|E_i \cap F_k) \text{ } (\because \text{ }E_i, F_k: \text{ independent})\nonumber \\
    =& q(1-q)^{j-i} \cdot q(1-q)^{k-j-1} \cdot \frac{B(k)-x}{B(k)-B(i)}
\end{align}
We can perform a similar computation for $k=m-1$. However, the probability of event $F_{m-1}$ is different from that of Equation~\eqref{eq:k_chosen} because the $(m-1)$-th level is always chosen by Algorithm~\ref{alg:quantization_mechanism}. Thus, we have
\begin{gather} \label{eq:last_chosen}
    \text{Pr}(F_{m-1}) = \text{Pr}(m-1: \text{ chosen}, \text{ }(j+1, \cdots, m-2): \text{ not chosen}) = (1-q)^{m-j-2}
\end{gather}
Therefore, for $k=m-1$, by combining Equations~(\ref{eq:i_chosen}), (\ref{eq:last_chosen}), and (\ref{eq:randomized_rounding}), we obtain
\begin{align} \label{eq:quantization_i_at_m-1}
    &\text{Pr}((Q(x)=i) \cap E_i \cap F_{m-1}) \nonumber \\
    =& \text{Pr}(E_i \cap F_{m-1}) \cdot \text{Pr}(Q(x)=i|E_i \cap F_{m-1}) \nonumber \\
    =&  \text{Pr}(E_i) \cdot \text{Pr}(F_{m-1}) \cdot \text{Pr}(Q(x)=i|E_i \cap F_{m-1}) \text{ } \nonumber \\
    =& q(1-q)^{j-i} \cdot (1-q)^{m-j-2} \cdot \frac{B(m-1)-x}{B(m-1)-B(i)}
\end{align}
Finally, by combining Equations~(\ref{eq:quantization_i_before_m-1}) and (\ref{eq:quantization_i_at_m-1}), we get
\begin{align} \label{eq:discrete_prob_I}
    &\text{Pr}(Q(x)=i) \nonumber \\
    =& \sum_{k=j+1}^{m-1} \text{Pr}((Q(x)=i) \cap E_i \cap F_k) \nonumber \\
    =& q(1-q)^{j-i}  \Big((1-q)^{m-j-2}\frac{B(m-1)-x}{B(m-1)-B(i)} + \sum\limits_{k=j+1}^{m-2} q(1-q)^{k-j-1} \frac{B(k)-x}{B(k)-B(i)}\Big) 
\end{align}

(II) $i=0$:

We can compute $\text{Pr}(Q(x)=0)$ in a similar way as in case (I). However, the probability of event $E_0$ is different from that of Equation~(\ref{eq:i_chosen}) because the $0$-th level is always chosen by Algorithm \ref{alg:quantization_mechanism}. Thus, we have 
\begin{gather} \label{eq:e_0}
    \text{Pr}(E_0) = \text{Pr}(0: \text{ chosen}, \text{ }(1, \cdots, j): \text{ not chosen}) = (1-q)^{j}
\end{gather}
Therefore, in (\ref{eq:discrete_prob_I}), by substituting $E_i$ into $E_0$, we get
\begin{align}
    &\text{Pr}(Q(x)=0) \nonumber \\
    =& \sum_{k=j+1}^{m-1} \text{Pr}((Q(x)=0) \cap E_0 \cap F_k) \nonumber \\
    =& (1-q)^{j}  \Big((1-q)^{m-j-2}\frac{B(m-1)-x}{B(m-1)-B(0)} + \sum\limits_{k=j+1}^{m-2} q(1-q)^{k-j-1} \frac{B(k)-x}{B(k)-B(0)}\Big) 
\end{align}

(III) $j+1 \leq i < m-1$:

For $i$ within this range, we can similarly compute $\text{Pr}(Q(x)=i)$ as in (I). To obtain $Q(x)=i$, the $i$-th discrete level must always be chosen while the $(j+1)$-th, $\cdots$, $(i-1)$-th discrete levels must not be chosen by the sub-sampling step of our algorithm. Thus, since $i \geq j+1$, the probability of $i$-th discrete level being used for randomized rounding can be expressed by using $F_i$ (refer to (\ref{eq:k_chosen})).
\begin{gather} \label{eq:i_chosen_III}
    \text{Pr}(F_i) = \text{Pr}(i: \text{ chosen}, \text{ }(j+1, \cdots, i-1): \text{ not chosen}) = q(1-q)^{i-j-1}
\end{gather}
Let us denote $k$ as an index of the just previous level less than $x$. The possible $k$'s are $0, \cdots, j$. Then, for $k \in [1, j]$, the probability of the $k$-th discrete level being used for randomized rounding can be represented by utilizing $E_k$ (refer to (\ref{eq:i_chosen})).
\begin{gather} \label{eq:k_chosen_III}
    \text{Pr}(E_k) = \text{Pr}(k: \text{ chosen}, \text{ }(k+1, \cdots, j): \text{ not chosen}) = q(1-q)^{j-k}
\end{gather}
Then, the likelihood of $x$ transitioning to the $i$-th discrete level due to the randomized rounding between $k$-th and $i$-th levels is $\frac{x-B(k)}{B(i)-B(k)}$. This means
\begin{gather} \label{eq:randomized_rounding_III}
    \text{Pr}(Q(x)=i|E_k \cap F_i) = \frac{x-B(k)}{B(i)-B(k)}
\end{gather}
Therefore, for $k \in [1, j]$, by combining (\ref{eq:i_chosen_III}), (\ref{eq:k_chosen_III}), (\ref{eq:randomized_rounding_III}), we get
\begin{align} \label{eq:quantization_i_after_0}
    &\text{Pr}((Q(x)=i) \cap E_k \cap F_i) \nonumber \\
    =&  \text{Pr}(F_i) \cdot \text{Pr}(E_k) \cdot \text{Pr}(Q(x)=i|E_k \cap F_i) \text{ } \nonumber \\
    =& q(1-q)^{i-j-1} \cdot q(1-q)^{j-k} \cdot \frac{x-B(k)}{B(i)-B(k)}
\end{align}
We can similarly calculate for $k=0$ by using (\ref{eq:e_0}).
\begin{align} \label{eq:quantization_i_at_0}
    &\text{Pr}((Q(x)=i) \cap E_0 \cap F_i) \nonumber \\
    =&  \text{Pr}(F_i) \cdot \text{Pr}(E_0) \cdot \text{Pr}(Q(x)=i|E_0 \cap F_i) \text{ } \nonumber \\
    =& q(1-q)^{i-j-1} \cdot (1-q)^{j} \cdot \frac{x-B(0)}{B(i)-B(0)}
\end{align}
Finally, by combining (\ref{eq:quantization_i_after_0}) and (\ref{eq:quantization_i_at_0})
\begin{align}
    &\text{Pr}(Q(x)=i) \nonumber \\
    =&\sum_{k=0}^{j} \text{Pr}((Q(x)=i) \cap E_k \cap F_i) \nonumber \\
    =&q(1-q)^{i-j-1}  \Big((1-q)^{j}\frac{x-B(0)}{B(i)-B(0)} + \sum\limits_{k=1}^{j} q(1-q)^{j-k} \frac{x-B(k)}{B(i)-B(k)}\Big)
\end{align}

(IV) $i=m-1$:

We can calculate $\text{Pr}(Q(x)=m-1)$ in a similar way compared to case (III). However, the ($m-1$)-th level should be always chosen by Algorithm \ref{alg:quantization_mechanism}, we rely on Equation~(\ref{eq:last_chosen}) rather than Equation~(\ref{eq:i_chosen_III}). We obtain:
\begin{align}
    &\text{Pr}(Q(x)=m-1) \nonumber \\
    =& \sum_{k=0}^j \text{Pr}((Q(x)=m-1) \cap E_k \cap F_{m-1}) \nonumber \\
    =& (1-q)^{m-j-2}  \Big((1-q)^{j}\frac{x-B(0)}{B(m-1)-B(0)} + \sum\limits_{k=1}^{j} q(1-q)^{j-k} \frac{x-B(k)}{B(m-1)-B(k)}\Big)
\end{align}
Therefore, we finally get Equation~(\ref{eq:discrete_prob_dist}) of Lemma \ref{lemma:discrete_prob_dist} from combining cases (I), (II), (III), and (IV).

\subsection{Proof of Theorem \ref{theorem:single_device}} \label{appx:proof_single_device}

We use Lemma~\ref{lemma:monotonicity} to find an upper bound on $D_\alpha (P_{Q(x)}||P_{Q(x^\prime)})$.
\begin{align*}
    D_\alpha (P_{Q(x)}||P_{Q(x^\prime)}) &\leq D_\infty (P_{Q(x)}||P_{Q(x^\prime)}) \\
    &\leq D_\infty (P_{Q(c)}||P_{Q(-c)}) \\
    &= \sup_{i \in \{0, 1, \cdots, m-1\}} \log \Big( \frac{P(Q(c)=i)}{P(Q(-c)=i)} \Big)
\end{align*}
The second inequality comes from the fact that the Renyi divergence is maximized at extreme points: this is because $(P, Q) \rightarrow D_\alpha (P||Q)$ is quasi-convex as observed in~\citep{van2014renyi}. We establish a value for $j$ that makes it so that $-c$ falls within the range of values between $B(j)$ and $B(j+1)$.  Since $P(Q(c)=i) \leq 1$ and $\arg\min_i P(Q(-c) = i) = m-1$, we obtain
\begin{align*}
    D_\alpha (P_{Q(x)}||P_{Q(x^\prime)}) &\leq \sup_{i \in \{0, 1, \cdots, m-1\}} \log \Big( \frac{P(Q(c)=i)}{P(Q(-c)=i)} \Big) \\
    &\leq \log \Big(\frac{1}{P(Q(-c) = m-1)} \Big) \\
    &= \log \bigg( \frac{1}{(1-q)^{m-2}\cdot\frac{-c-B(0)}{B(m-1)-B(0)} + \sum\limits_{k=1}^j q(1-q)^{m-2-k}\cdot\frac{-c-B(k)}{B(m-1)-B(k)}} \bigg) \\
    &= \log \bigg( \frac{1}{(1-q)^{m-2}\cdot\frac{\Delta}{2c+2\Delta} + \sum\limits_{k=1}^j q(1-q)^{m-2-k}\cdot\frac{-c-B(k)}{c+\Delta-B(k)}} \bigg) \\
    &\leq \log \frac{1}{(1-q)^{m-2}\cdot \frac{\Delta}{2c+2\Delta}} \\
    &= \log \left(\frac{2(1-q)^2(c+\Delta)}{\Delta}\right) + m\log\frac{1}{1-q}
\end{align*}
To go from the second to the third line, we used Lemma \ref{lemma:discrete_prob_dist}.

\section{Experimental setup details} \label{appx:experimental_setup_details}

\paragraph{Implementation environment.} We adopt the same implementation setup as outlined in \citet{chen2022poisson}. To implement our algorithm, we utilize TensorFlow \citep{tensorflow2015-whitepaper} and the TensorFlow Federated (TFF) library. Our computational resources include 2 NVIDIA RTX A5000 GPUs. We simulate a federated learning scenario involving a total of $3,400$ local devices, with $n = 40$ local devices participating in each round
. The total number of communication rounds is set to $2,000$.

\paragraph{Dataset \& training model.} We perform image classification on the EMNIST dataset, which is comprised of 62 classes. We employ a Convolutional Neural Network (CNN) as the learning model for our training purposes.

\section{Additional experimental results} \label{appx:more_experiments}

In Section \ref{sec:experiments}, we opted for $\theta = 0.25$ as a hyperparameter for PBM and identified the pairs $(\Delta, q)$ - hyperparameters for RQM - that yielded a superior privacy-accuracy trade-off. Here, in this section, we extend our investigation by conducting additional experiments with a range of $\theta$ values within the $[0, 0.4]$ interval, seeking to underscore the 
superiority of RQM in achieving a better privacy-accuracy trade-off compared to PBM across a wider range of $\theta$ values considered by~\citet{chen2022poisson}. For these subsequent trials, we choose $\theta = 0.15$ and $\theta = 0.35$.

\subsection{Additional DP experiments} \label{appx:more_dp_experiments}

In line with the experimental protocol detailed in Section \ref{sec:dp_experiment_result}, we conduct additional trials using differing $\theta$ values: $\theta = 0.15$ and $\theta = 0.35$. Consistent with the results reported in \ref{sec:dp_experiment_result}, these extended trials again demonstrate RQM's superior performance, reflected in its lower Renyi divergence, and thus enhanced privacy guarantee when compared with PBM.

\begin{figure*}[hbt!]
    \centering
    \begin{subfigure}[b]{0.4\textwidth}
    \includegraphics[width=\textwidth]{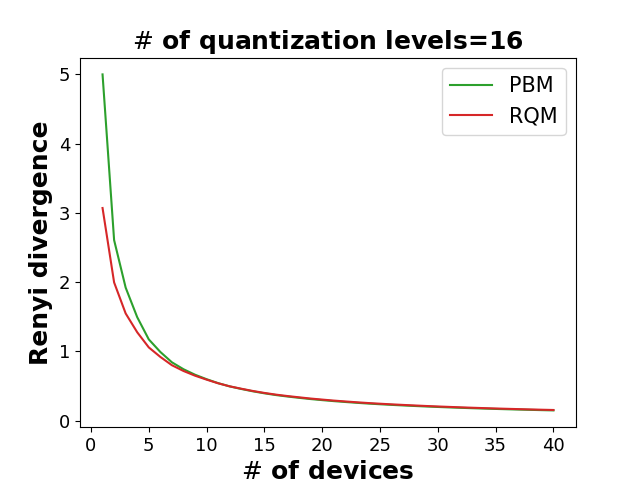}
    \end{subfigure}
    \quad
    \begin{subfigure}[b]{0.4\textwidth}
    \includegraphics[width=\textwidth]{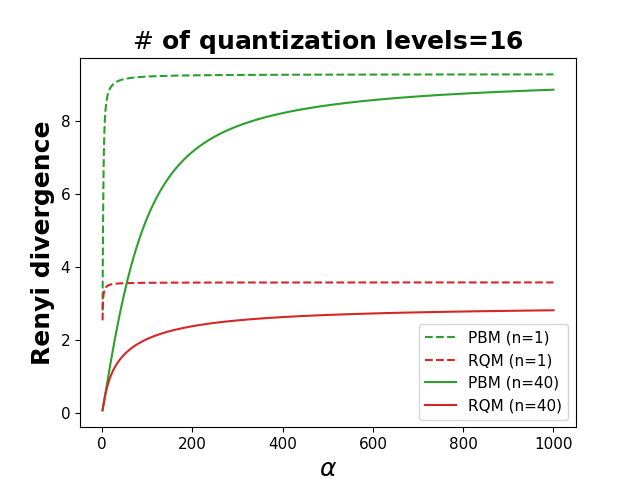}
    \end{subfigure}
    \caption{The results of an additional experiment about Numerical Renyi privacy guarantees with $\theta = 0.15$.}
    \label{fig:renyi_devices_alpha_scale_15}
\end{figure*}

\paragraph{Additional experiment with $\theta = 0.15$.} We choose $(\Delta, q) = (2.33c, 0.42)$ for RQM to compare with PBM with $\theta = 0.15$. The results of this experiment can be found in Figure \ref{fig:renyi_devices_alpha_scale_15}.

\begin{figure*}[hbt!]
    \centering
    \begin{subfigure}[b]{0.4\textwidth}
    \includegraphics[width=\textwidth]{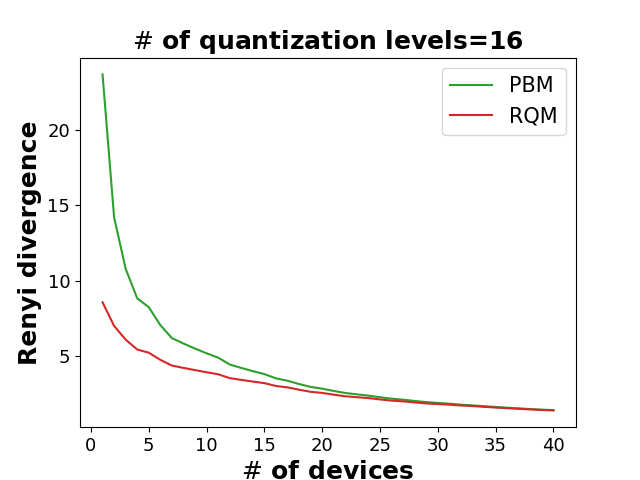}
    \end{subfigure}
    \quad
    \begin{subfigure}[b]{0.4\textwidth}
    \includegraphics[width=\textwidth]{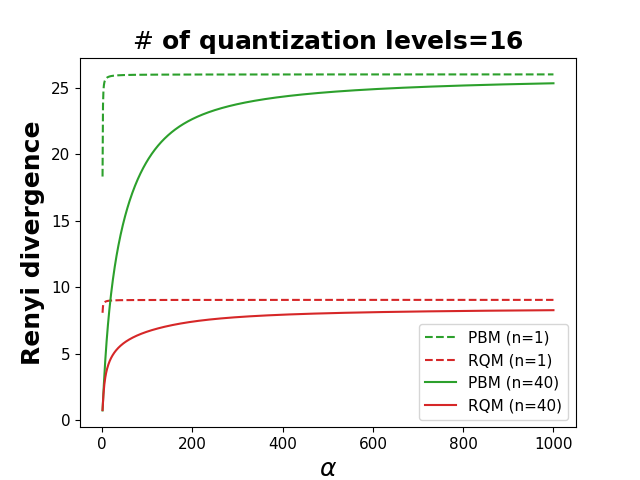}
    \end{subfigure}
    \caption{The results of an additional experiment about Numerical Renyi privacy guarantees with $\theta = 0.35$.}
    \label{fig:renyi_devices_alpha_scale_35}
\end{figure*}

\paragraph{Additional experiment with $\theta = 0.35$.} We choose $(\Delta, q) = (0.429c, 0.49)$ for RQM to compare with PBM with $\theta = 0.35$. The results of this experiment can be found in Figure \ref{fig:renyi_devices_alpha_scale_35}.

\subsection{Additional FL experiments}

We follow the same FL experimental setup as in Section \ref{sec:fl_experiment}. For the choice of hyperparameters, we use the same hyperparameters for the additional FL experiments as those of Appx.~\ref{appx:more_dp_experiments}. We also plot results for two more $(\Delta, q)$ pairs, for both $\theta$ values $\theta=0.15$ and $\theta=0.35$. Echoing the findings of Section \ref{sec:fl_experiment}, our additional analysis shows that, for the above values of $\theta$, all three instantiations of RQM consistently outperform PBM in terms of both performance (measured both in terms of loss and accuracy) and Renyi DP guarantees.

\paragraph{Additional experiment with $\theta = 0.15$.} Other than $(\Delta, q) = (2.33c, 0.42)$, we also plot additional results for $(\Delta, q) = (4c, 0.5)$ and $(\Delta, q) = (c, 0.23)$. All three RQMs achieve similar accuracy, which is higher than one achieved by PBM with $\theta=0.15$.

\begin{figure*}[hbt!]
    \centering
    \begin{subfigure}[b]{0.32\textwidth}
    \includegraphics[width=\textwidth]{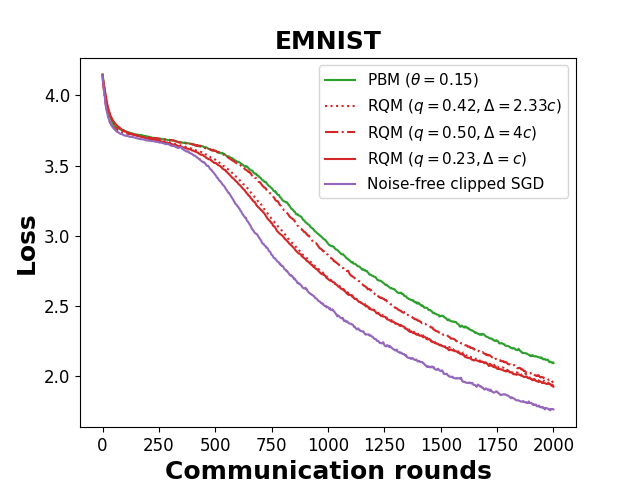}
    \end{subfigure}
    \quad
    \begin{subfigure}[b]{0.32\textwidth}
    \includegraphics[width=\textwidth]{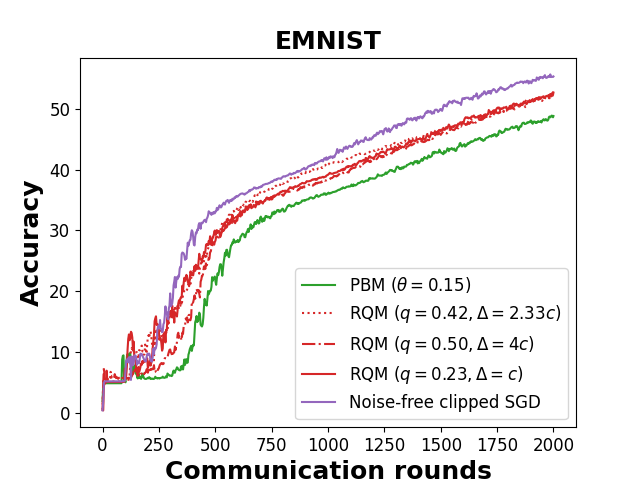}
    \end{subfigure}
    \begin{subfigure}[b]{0.32\textwidth}
    \includegraphics[width=\textwidth]{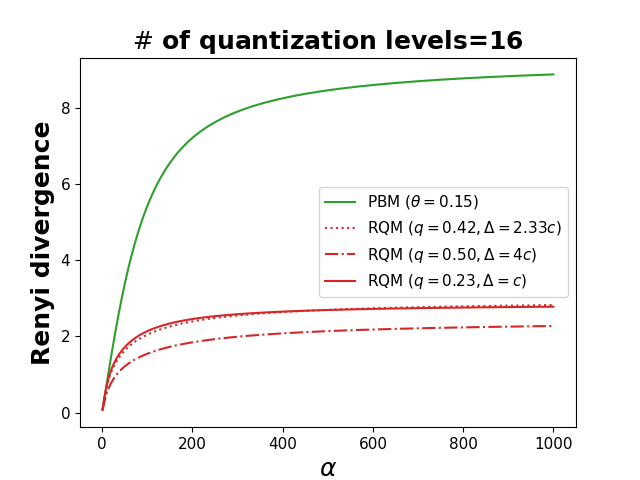}
    \end{subfigure}
    \caption{Comparing RQM with PBM and noise-free clipped SGD on EMNIST (Additional FL experiment with $\theta=0.15$).}
    \label{fig:emnist_fl_scale_15}
\end{figure*}

\paragraph{Additional experiment with $\theta = 0.35$.} The total number of communication rounds for this additional experiment is $1700$. For the hyperparameters of RQM, other than $(\Delta, q) = (0.429c, 0.49)$, we also plot results of two more pairs $(\Delta, q) = (c, 0.65)$ and $(\Delta, q) = (0.25c, 0.37)$. All three RQMs achieve higher accuracy than one achieved by PBM with $\theta=0.35$. Among three RQMs, $(\Delta, q) = (c, 0.65)$ achieves the best accuracy. 

\begin{figure*}[hbt!]
    \centering
    \begin{subfigure}[b]{0.32\textwidth}
    \includegraphics[width=\textwidth]{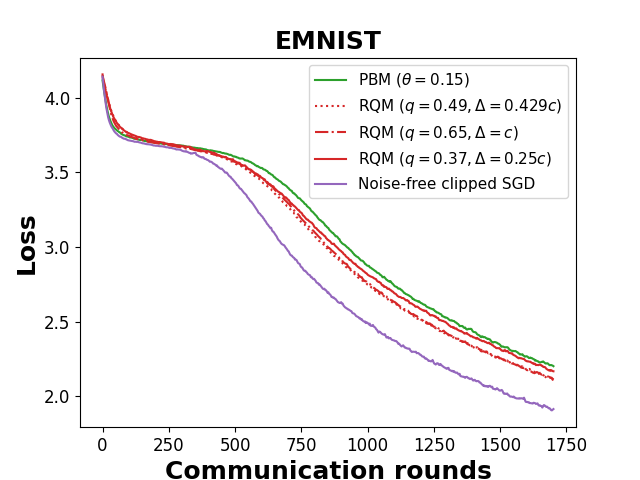}
    \end{subfigure}
    \quad
    \begin{subfigure}[b]{0.32\textwidth}
    \includegraphics[width=\textwidth]{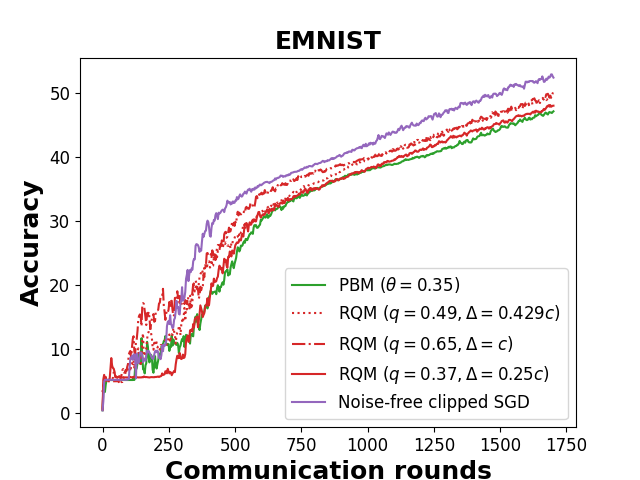}
    \end{subfigure}
    \begin{subfigure}[b]{0.32\textwidth}
    \includegraphics[width=\textwidth]{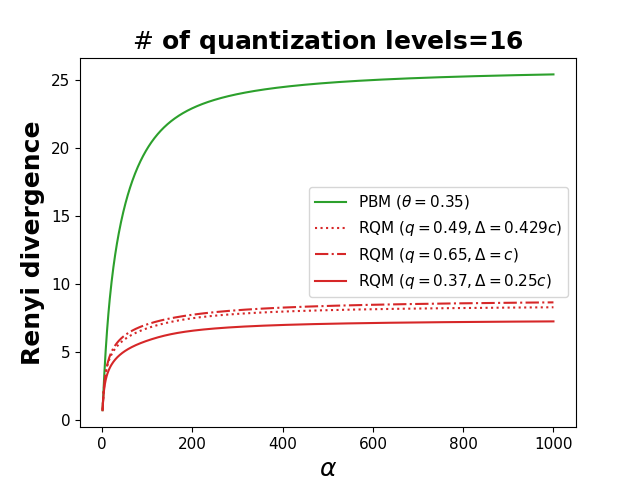}
    \end{subfigure}
    \caption{Comparing RQM with PBM and noise-free clipped SGD on EMNIST (Additional FL experiment with $\theta=0.35$).}
    \label{fig:emnist_fl_scale_35}
\end{figure*}

\end{document}